\documentclass[10pt,twocolumn,letterpaper]{article}

\usepackage{cvpr}              

\usepackage{graphicx}
\usepackage{amsmath}
\usepackage{amssymb}
\usepackage{booktabs}
\usepackage{comment}
\usepackage{graphics}
\usepackage{amsmath}

\usepackage[pagebackref,breaklinks,colorlinks]{hyperref}

\usepackage[capitalize]{cleveref}
\crefname{section}{Sec.}{Secs.}
\Crefname{section}{Section}{Sections}
\Crefname{table}{Table}{Tables}
\crefname{table}{Tab.}{Tabs.}

\def\cvprPaperID{1571} 
\def\confName{CVPR}
\def\confYear{2022}

\newcommand{\nerf}{NeRF\@\xspace}
\newcommand{\nerfs}{NeRFs\@\xspace}

\newcommand{\bfe}{\mathbf f}
\newcommand{\bn}{\mathbf n}
\newcommand{\bo}{\mathbf o}

\newcommand{\bmu}{\boldsymbol\mu}
\newcommand{\cP}{{\cal P}}

\begin{document}

\title{ONeRF: Unsupervised 3D Object Segmentation from Multiple Views}

\author{
\begin{tabular}{ccccc}
Shengnan Liang$^{1}$\thanks{ Equal contribution. The order of authorship was determined alphabetically. } & 
Yichen Liu$^{1}${\footnotemark[1]} & 
Shangzhe Wu$^{3}$ & 
Yu-Wing Tai$^{1,2}$ &
Chi-Keung Tang$^{1}$ 
\end{tabular}
\\
\begin{tabular}{ccc}
$^1$The Hong Kong University of Science and Technology &
$^2$Kuaishou Technology &
$^3$University of Oxford
\end{tabular} \\
}
\maketitle

\begin{abstract}
We present ONeRF, a method that automatically segments and reconstructs object instances in 3D from multi-view RGB images without any additional manual annotations.
The segmented 3D objects are represented using separate Neural Radiance Fields (NeRFs) which allow for various 3D scene editing and novel view rendering.
At the core of our method is an unsupervised approach using the iterative Expectation-Maximization algorithm, which effectively aggregates 2D visual features and the corresponding 3D cues from multi-views for joint 3D object segmentation and reconstruction.
Unlike existing approaches that can only handle simple objects, our method produces segmented full 3D NeRFs of individual objects with complex shapes, topologies and appearance. The segmented O\nerfs enable a range of 3D scene editing, such as object transformation, insertion and deletion. 
\end{abstract}

\section{Introduction}

Neural Radiance Fields (NeRF)~\cite{mildenhall2020nerf} have recently been becoming the mainstream approach for novel view synthesis, given its excellent performance for complex real-world scenes despite their simplicity.
The key idea is to represent the entire scene as a radiance field parametrized using an MLP, taking in $xyz$ coordinates and viewing directions as input and produces densities and view-dependent colors.
Since the first contribution~\cite{mildenhall2020nerf}, NeRfs have been extended in various dimensions, for example, with improved efficiency~\cite{yu2021plenoctrees,kangle2021dsnerf},
improved quality~\cite{kaizhang2020,Barron_2021_ICCV},
deformation~\cite{park2021nerfies,Pumarola_2021_CVPR},
uncalibrated images~\cite{wang2021nerfmm,SCNeRF2021},
sparse views~\cite{Jain_2021_ICCV,Chibane_2021_CVPR},
better generalization~\cite{Schwarz2020NEURIPS,Yu_2021_CVPR},
material decomposition~\cite{boss2021nerd,zhang2021nerfactor},
as well as many other applications~\cite{Guo_2021_ICCV,Liu_2021_ICCV}.



\begin{figure}[t]
    \includegraphics[width=1\linewidth]{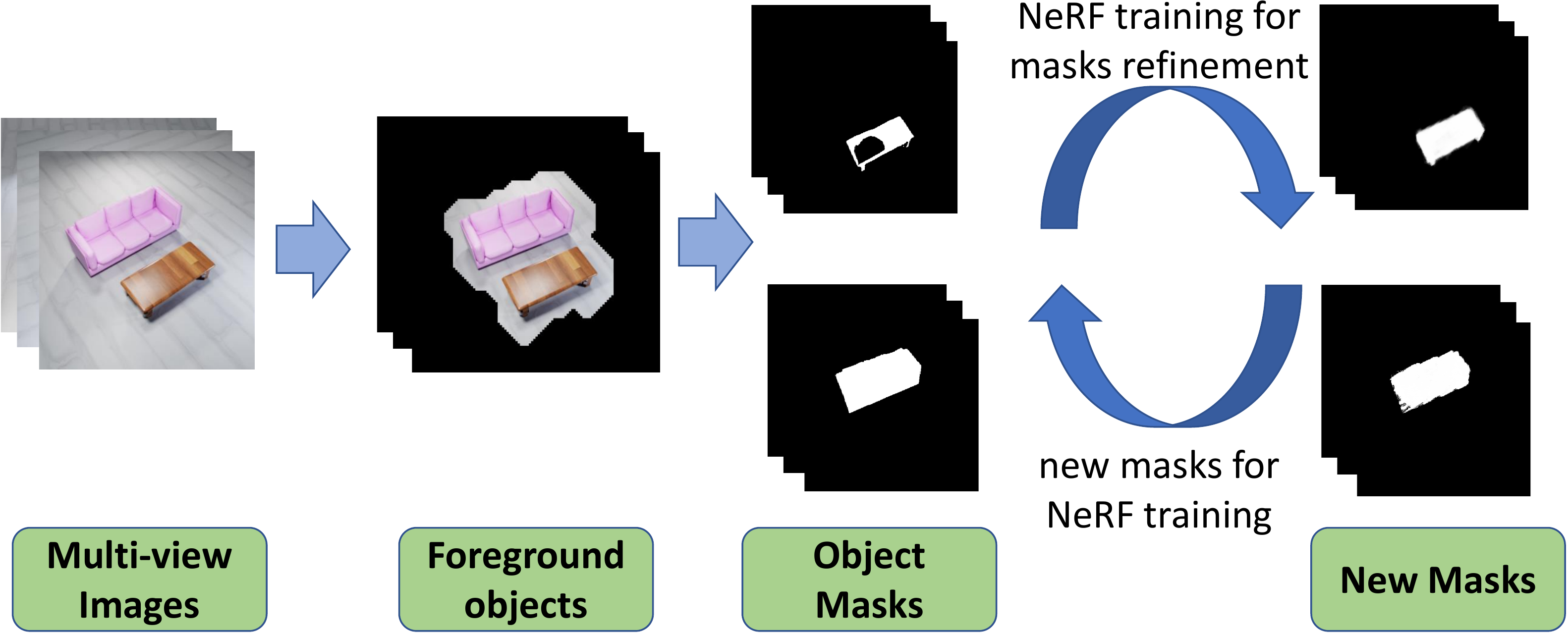}
    \caption{\textbf{3D slot attention.} Coarse 2D masks are iteratively refined with 3D object \nerfs in an unsupervised manner. The output are individual \nerfs for each object.}
    \label{fig:teaser}
\end{figure}

However, most of the existing work focuses on novel view rendering. 
Less effort has been made to leverage this representation for the fundamental computer vision problem of 3D object segmentation, which is the goal of this paper.
Notably, by obtaining object-level \nerf representations of a 3D scene, various 3D scene editing tasks can be enabled, including object editing, insertion, removal and completion.
Several recent works have attempted to learn category-specific object \nerfs from multi-view images and optional depth information, allowing for single-image inference~\cite{stelzner2021,yu2021}.
However, these methods mainly work with specific object categories, such as CLEVR~\cite{Johnson_2017_CVPR} and ShapeNet~\cite{Wu_2015_CVPR} objects, and is not capable of segmenting arbitrary objects in general scenes.
Others rely on additional information, such as 2D segmentation~\cite{yang2021objectnerf} and 2D bounding box annotations~\cite{zhang2021stnerf}, and lift them to 3D.
NeuralDiff~\cite{tschernezki21neuraldiff} on the other hand leverages motion cues in egocentric videos to separate the foreground layer of moving objects and the static background layer, but cannot obtain full 3D segmentations of individual foreground objects.
In contrast, our O\nerfs consist of an unsupervised approach that performs automatic 3D object segmentation of arbitrary scenes from multiple views, without relying on any external annotations, which can be regarded as a generalized version of some of the above category-specific methods.

Given multi-view images of a scene, our goal is to detect individual objects and represent each one of them using a separate NeRF. In order to do this, 
we introduce the notion of {\em 3D slot attention}, which is an unsupervised method for fusing 2D visual features and 3D geometrical information inherent from multiple views  
for segmenting the scene into individual 3D object \nerfs. 
Specifically, both the 2D features and 3D information extracted from  multiple views are used  
to produce initial object and background masks, which will be iteratively refined via expectation-maximization to segment individual \nerfs and thus 3D objects. Unlike previous works, our unsupervised O\nerfs segmentation can handle objects with complex textures, geometries and topologies, which support not only novel view synthesis but also scene editing such as object insertion, deletion, and modification with individual O\nerfs automatically learned from a given scene. 

While the qualitative results have significant room for improvements as in other related state-of-the-arts, this paper makes the first attempt to contribute a significant baseline for unsupervised learning on segmenting a cluttered scene into individual O\nerfs from multiple views. Our qualitative and quantitative results have validated  the soundness of our technical approach. Preliminary results on scene editing shows the great potential of O\nerfs.

\section{Related Work}

\subsection{Multi-View 3D Reconstruction}
Multi-view stereo (MVS) is a classical computer vision problem for 3D reconstruction from images captured from
multiple viewpoints.  
Traditional methods include~\cite{Bonet_poxels:probabilistic,
Kutulakos00atheory,
kolmogorov2002multi-camera,
pmvs_pami}. More recent deep learning-based MVS methods are MVSNet~\cite{yao2018mvsnet}, P-MVSNet~\cite{Luo_2019_ICCV}, and
CVP-MVSNet~\cite{Yang_2020_CVPR}.
Once reconstructed, the 3D scene can be rendered at any viewpoints. However, the output of MVS consists of (quasi) dense point clouds and thus are not object aware.
Combining deep MVS technique with differentiable volume rendering, MVSNeRF~\cite{chen_mvsnerf} reconstructs a neural radiance field from unstructured multi-view input images for novel view synthesis, which generalizes well across scenes using MVS. View synthesis from deep lightfield~\cite{LearningViewSynthesis} has also been proposed. While novel views can be synthesized, the pertinent \nerf has little object awareness.

\subsection{Object Segmentation}

In our unsupervised approach,
2D object masks estimated from multiple views must be consistent with the underlying 3D voxel volume. 3D visual hulls~\cite{ibvh} can be estimated from accurate 2D object masks given in the input calibrated images. Conversely, accurate
segmentation of 3D point clouds/voxels can produce consistent 2D object masks across views. Segmentation of 3D point cloud/voxels is an active research area, including
PointNet~\cite{Qi_2017_CVPR},
DGCNN~\cite{dgcnn} and
\cite{qiu2021geometric}, and later on large-scale point cloud segmentation such as
PointNet++~\cite{qi2017pointnetplusplus} and
~\cite{Liu_2019_CVPR,Hu_2020_CVPR,Yan_2020_CVPR}.
Recent works include weakly supervised segmentation~\cite{Zhang_2021_ICCV,Zhou_2021_ICCV},
semi-supervised segmentation~\cite{Jiang_2021_ICCV},
few shot segmentation~\cite{Zhao_2021_CVPR},
instance segmentation~\cite{Zhang_2021_CVPR}, and
semantic segmetation~\cite{Qiu_2021_CVPR}. While the above can be deployed for 3D \nerf segmentation on the pertinent voxel volume, ONeRFs is an end-to-end unsupervised approach for object \nerfs  segmentation taking only images as input.

There is also another line of work on unsupervised object-centric learning, which is learning-based and decomposes a single image into different objects, mostly in 2D: e.g., GENISIS~\cite{genesis}, SPACE~\cite{SPACE}, SPAIR~\cite{crawford2019spatially}, IODINE~\cite{IODINE}), SlotAttention~\cite{SlotAttention}. Most of these work with very simple CLEVR-like objects.

\begin{figure*}[t]
    \includegraphics[width=1\linewidth]{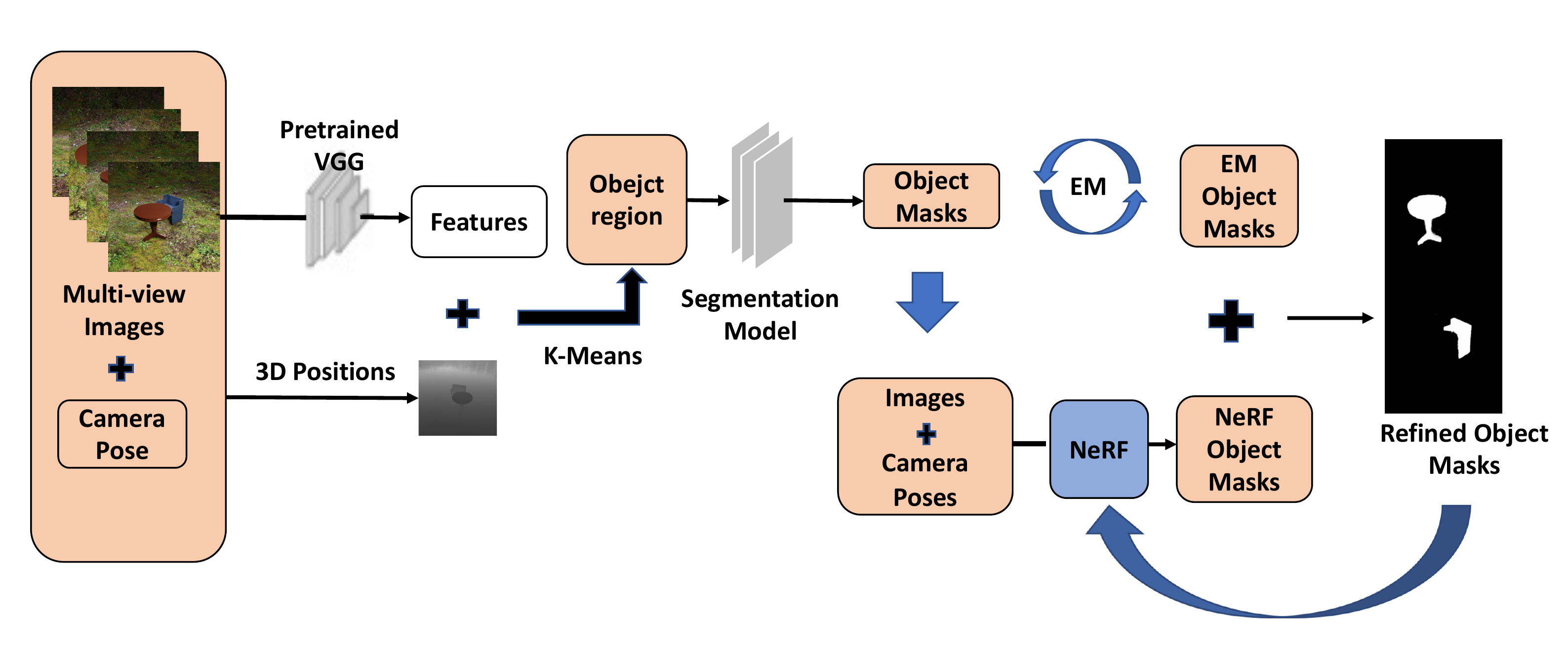}
    \caption{\textbf{Overall framework.} 
    Given a set of images of a scene and their camera poses, we use K-Means to obtain coarse foreground object regions according to the VGG feature and 3D position of each pixel. Then we obtain the object masks by using an unsupervised object segmentation convolutional neural network, and then use these masks to train NeRF networks for individual objects. Combining the masks generated by NeRF and the masks improved by expectation maximization (EM), we get the refined objects masks, which will be used to train NeRF models in next iteration.
    }
    \label{fig:pipeline}
\end{figure*}

\section{Method}
ONeRFs is an end-to-end unsupervised approach for segmenting 3D object NeRFs from multiple views of the underlying scene. Borrowing attention typical of deep neural networks, our 3D attention slots operates by cooperating 2D visual and 3D geometrical cues, so that they can reinforce each other in  alternating optimization where respective errors will be automatically corrected via resolving their inconsistencies. Figure~\ref{fig:pipeline} shows the overall framework, where starting coarse 2D masks are working in tandem with 3D reconstruction from \nerfs to iteratively achieve accurate object \nerfs segmentation. 

\subsection{Initial coarse masks generation}

\noindent {\bf Clustering.}
    Given the input images of the scene and camera poses from multiple viewpoints, we resize the images to 128 $\times$ 128 and extract VGG features~\cite{Simonyan15} and  reconstruct 3D positions of all the pixels, which are clustered into two classes using $K$-Means clustering: foreground and background.
    Let
    \begin{eqnarray}
        s^f_i &=& ||\bfe_i-\bmu_f||+w||P_i-\bold p_f||,\\
        s^b_i &=& ||\bfe_i-\bmu_b||,
    \end{eqnarray}
where $s_i^f$ and $s_i^b$ are respectively the foreground and background scores of pixel $i$, $\bfe_i$ is the VGG feature,
    $P_i$ are the 3D coordinates of corresponding pixel $i$, which can be obtained from multi-view reconstruction from the input images. $\bmu_f$ and $\bmu_b$ are respectively the mean foreground and background clustered features, $\bold p_f$ are the 3D coordinates corresponding to the clustered foreground points, and $w$ is the weight. 
    
Assuming  sufficiently far background, background pixels will have large variance in 3D locations. Thus foreground pixels tend to have smaller $||P_i-\mu_f||$. Then we update the three means by 
    \begin{eqnarray}
        \bmu_f &=& \frac{\sum_{i:s_i^f<s_i^b}\bfe_i}{\#},
        \bold p_f = \frac{\sum_{i:s_i^f<s_i^b}P_i}{\#},\\
        \bmu_b &=& \frac{\sum_{i:s_i^f\ge s_i^b}\bfe_i}{\#},
    \end{eqnarray}
where \#'s correspond to the relevant number of foreground and background pixels. 
Finally we get the foreground and background masks $M_f^0$, $M_b^0$, and upsample them into the original size.


\noindent {\bf Image Segmentation and Instances.} After labeling the  pixels corresponding to the 3D foreground clustered above, we apply unsupervised segmentation~\cite{9151332} to produce $n+1$ parts 
$\cP_0, \cdots, \cP_{n}$. 
$\cP_0$ is the background we separate in the first step, $\cP_1$ is the background near the foreground objects which cannot be precisely clustered in the first step and the others correspond to the foreground objects, denoted as $\cP_2,\cdots,\cP_n$.

Suppose there are $k$ objects in expectation. We then discard $n-(k+1)$ of $\cP_1, \cdots \cP_n$ occupying the smallest areas, i.e. those classes with fewest pixels, which are very likely due to incorrect clustering or noise.

We then put the full image into the model in image segmentation step to obtain the response vector map of the image~\cite{9151332}, where the response vector at a given pixel represents the clustering scores for all the $n$ classes (minus the weak ones rejected as above). The pixels in the same class tend to have similar response vectors. Here,  we cluster the pixels by the highest entry in  the response vector.
The response vectors of pixels in $\cP_0$ are similar
to those in $\cP_1$. We sample some pixels in $\cP_i$ and find the mean of the response vectors of those pixels, denoted as $\bn_i$. Then, $\cP_i$ with $\bn_i$ that is closest to $\bn_0$ is the remaining background to be separated from the coarse foreground.

\subsection{Train NeRFs and Refine Masks}
\noindent {\bf Training.}
Given a 3D position $\bold x$ and view direction $\bold d$, the network will output the RGB color $\bold c$ and volume density $\sigma$ of that point~\cite{mildenhall2020nerf}.
Following~\cite{yang2021objectnerf},
%
for each segmented object $a$ the loss is
\begin{equation}
\lambda_1||\hat C_a(\bold r)-C(\bold r)|| \odot M_a +\lambda_2||\hat A_a(\bold r)-M_a||
\end{equation}
where $\bold r$ is the ray, $\hat C_a(\bold r)$ is the RGB triplet by the \nerf of object $a$, \(C(\bold r)\)
is the ground truth RGB, $M_a$ is the mask of object $a$, and
$\hat A_a(\bold r)$ is the accuracy map by NeRF of object $a$, and $\lambda_1, \lambda_2$ are weights. $C_a(\bold r)$ and $A_a(\bold r)$, which follows ~\cite{mildenhall2020nerf}, are defined as 
\begin{eqnarray}
\hat C_a(\bold r) &=& \sum_{i=1}^N T_i^a\alpha_i^a \bold c_i^a
\label{eq:c_a}  \\
\hat A_a(\bold r) &=& \sum_{i=1}^N T_i^a\alpha_i^a \label{eq:a_a} \\
T_i^a &=& \sum_{j=1}^i \exp(-\sigma_j^a\delta_j)\\
\alpha_i^a &=& 1-\exp(-\sigma_j^a\delta_j)
\end{eqnarray}
We sample $N$ points along the ray $\bold r$ and the distance between the $i$-th  point and the camera is $\delta_i$, which will increase with the increase of the index. For point $i$ in object~$a$, we obtain the RGB color $\bold c_i^a$ and volume density $\bold \sigma_i^a$ at that point from the network. Then by Equation~(\ref{eq:c_a}) and~(\ref{eq:a_a}), we obtain the rendered color and accuracy masks along that ray.

The color within the in-mask region will converge to the ground-truth RGB color. For $\hat A_a$, in-mask region will converge to 1 while regions outside to 0.
As NeRF needs to guarantee 3D consistency, occlusion and incorrect masking in the last step can be handled automatically.

\noindent {\bf Refine Masks.}
The masks are iteratively refined by leveraging the object \nerfs
for each individual object instance, in the presence of occlusion if any. We use the occlusion accuracy masks $\hat A_a'(\bold r)$ as the occlusion masks from NeRF networks. Specifically, each new object mask is generated by
\begin{align}\hat A_a'(\bold r) &= \sum_{i=1}^N T_i\alpha_i^a\\
T_i &= \sum_{j=1}^i \exp(-\sigma_j\delta_j)\\
\sigma_j&=\sum_o \sigma_j^o
\end{align}
where $\sigma_j$ is the summation of volume densities from NeRF networks of all objects.
We only consider  the contribution of the other NeRFs on $T_i$, that is, decreasing values of $\alpha$ along the ray $\bold r$. Finally, expectation-maximization (EM) is employed to 
refine the input mask. 
We use the response vector map again. We
initialize the $\mu$ as

\begin{align}
\bold \mu_i^{(0)} =  \frac{\sum_{x_j\in M_i} \bold v(\bold x_j)}{\#}
\end{align}
where $M_i$ is the mask, $\bold v_j$ is the response vector and  \#'s correspond to the number of pixels in $M_i$. 
Then EM is performed on the pixels of background. At the $t$-th iteration, for each response vector of background $\bold v_n$, we formulate the posterior probability of $\bold v_n$ given $\bmu_i$ as
\begin{align}
p(\bold v_n|\bold \bmu_i^{(t)})=\exp(-||\bold v_n - \bmu||_2^2),
\end{align}
where $\bmu_i^{(t)}$ is the $\bmu$ value at the $t$-th iteration.
The latent variable $\bold z_n^{(t)}$ for $\bold v_n$ is
\begin{align}
z_{n,i}^{(t)} = \frac{p(\bold v_n|\bmu_i)}{\sum_{k=1}^K p(\bold v_n|\bmu_k)},
\end{align}
where $z_{n,i}^{(t)}$ is the $i$-th entry of $\bold z_n^{(t)}$ and $K$ is total number of classes, which is the number of objects plus one (the background). 
Then we update the $\bmu$ value of the next iteration as 
\begin{align}
\bold \bmu_{i}^{(t+1)} = \frac{\sum_{n=1}^Nz_{n,i} \bold v_n}{\sum_{n=1}^N z_{n,i}}
\end{align}
Then we output the softmax results upon the above EM convergence. Let $M_{re}$ be the resulting mask, then
the final mask is given by 
\begin{equation}
w_1M_{re}+w_2\hat A_a' > 0.5
\end{equation}
for some weights $w_1, w_2$.

\section{Experiments}

\subsection{Setup}
\subsubsection{Datasets}
We first test our method on CLEVR dataset~\cite{Johnson_2017_CVPR} and then move to more complex scenes. Our dataset is built upon ClevrTex~\cite{karazija2021clevrtex}, which wraps some realistic textures to the objects and background. Instead of regular geometric models,  we use the some objects with more realistic shapes, such as animals and furniture items, and use Blender~\cite{blender} to render $400 \times 400$ images from 30 different viewpoints for each scene, keeping the camera-to-world matrices for all such rendered images. Additionally, we render the ground truth depth map for evaluating the predicted 3D position. Each scene contains 2 to 4 objects with different colors. We use 25 images from segmentation and training and 5 for testing randomly. We test on 8 different scenes in total.

\subsubsection{Implementation Details}
To obtain the groundtruth 3D positions, we train a NeRF model~\cite{mildenhall2020nerf} for each scene and generate the depth map from each training viewpoints. 
By  combining camera rays and the depth maps, we obtain the corresponding 3D positions of the pixels.

We use the features of the first 16 layers of pretrained VGG models and upsample the feature maps to the original image size for clustering. After clustering, we dilate the object masks to reduce the mis-clustering of object pixels. For object segmentation, we train a segmentation network for each scene with batch size 10.

\begin{figure*}[t]
    \includegraphics[width=1\linewidth]{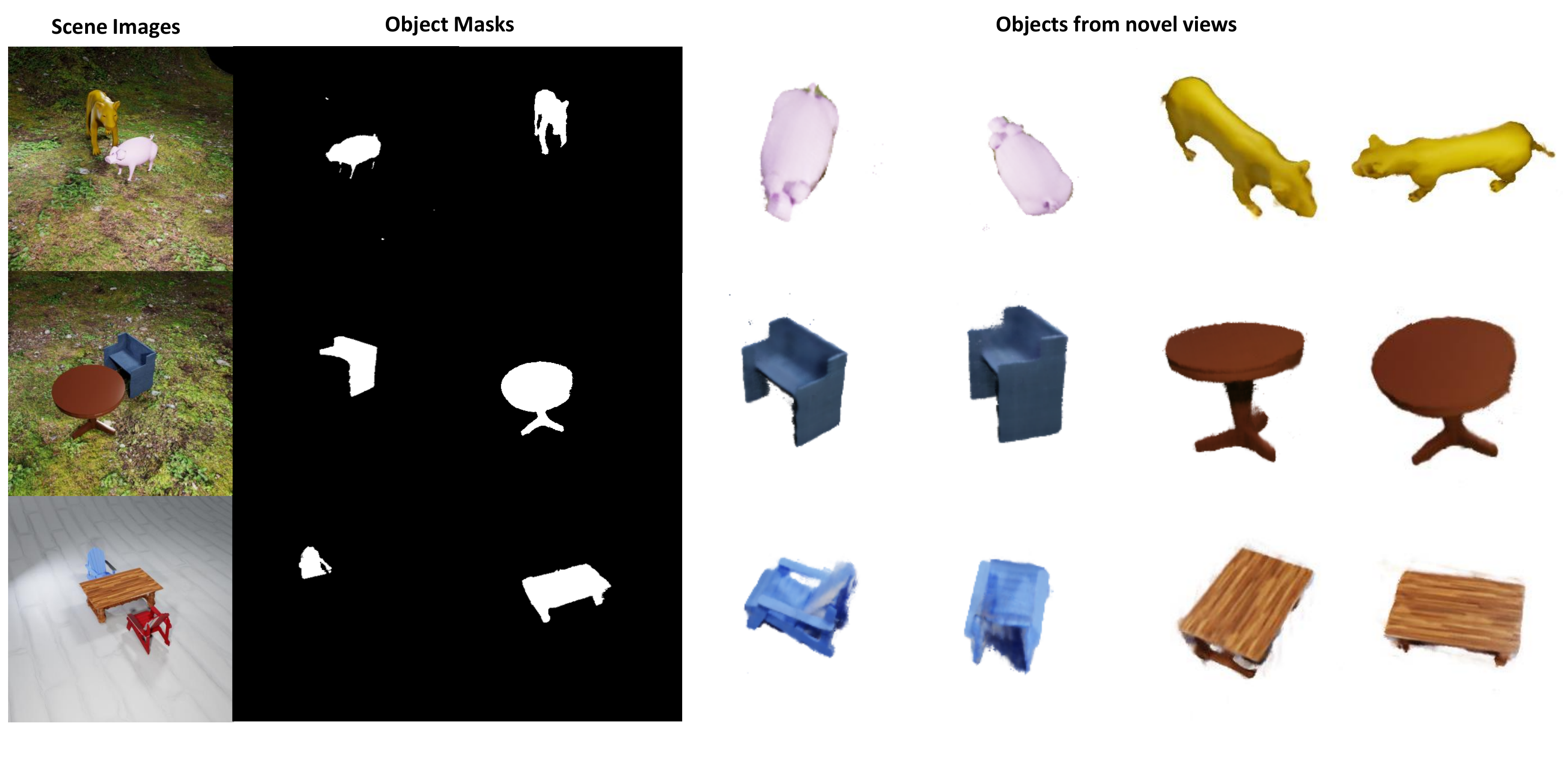}
    \caption{\textbf{Qualitative results.} This figure shows a sample input image for each of the three examples, refined masks of 
    the 3D objects in the scene, and output images of different objects from two testing views of each scene. The \nerfs are trained with 25 images. 
    }
    \label{fig:results}
\end{figure*}

\subsection{Qualitative Results}

\subsubsection{Qualitative Comparisons}

Figure~\ref{fig:results} shows qualitative results of some objects from unseen novel views. Specifically, to visualize the 3D segmentation and each object space, we render the segmented object \nerfs at novel viewpoints and obtain the final 2D masks. Notably, all novel views are rendered {\em without} using any masks, which validates the converged masks output by our method. 

Our method is scene-specific and unsupervised for 3D object segmentation. Notably, this approach is not learning-based and does not need large sets of data on similar scene for training. To our knowledge there is no non-learning-based methods for 3D object segmentation on \nerfs similar to ours. 
With the 3D information from \nerfs, we can obtain two kinds of masks that will be useful for further applications: 1) the mask for a certain object in original space; 2) the mask for certain object in individual object space. The results show that our method could handle complex textures (e.g., wood coffee table), and complex shapes (e.g., animals) and non-trivial topologies (e.g., 
genus-$n$ objects such as the chairs). 

\begin{figure*}[h]
\centering
    \includegraphics[width=0.99\linewidth]{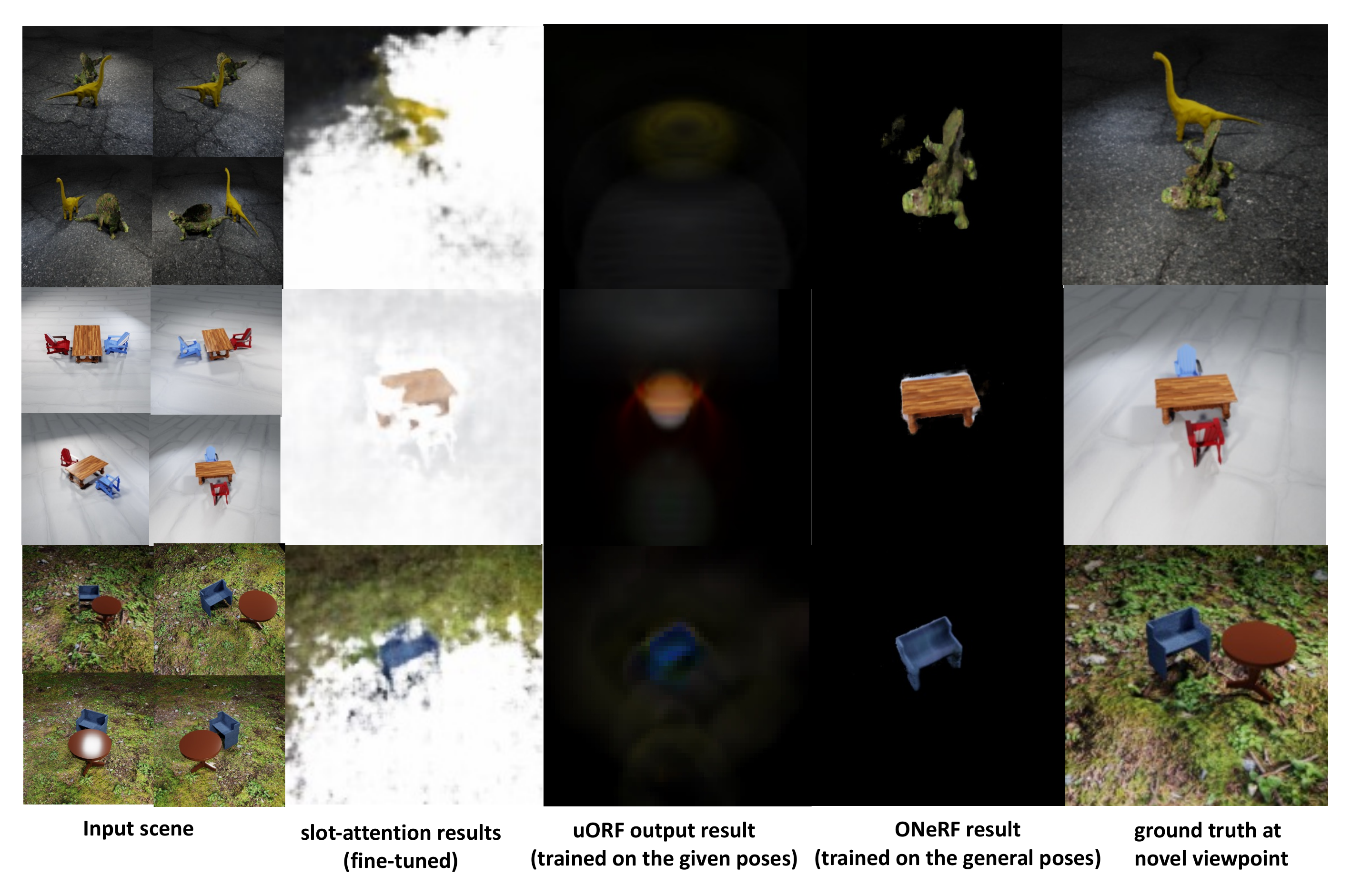}
    \caption{\textbf{Qualitative comparison.} Comparison with uORF~\cite{yu2021} and slot attention~\cite{SlotAttention}. The results show that our model is the most robust on the scenes with complex shape and color. Learning based methods which rely on large number of scenes with similar color and shape distribution is difficult to generalize on our dataset. 
    }
    \label{fig:uorf}
\end{figure*}

We compare our results with both 2D unsupervised segmentation methods and 3D learning-based unsupervised segmentation approaches. We cannot obtain reasonable results by directly applying learning-based methods such as uORF~\cite{yu2021} on our dataset. In our tests, we use the random camera poses but all the objects are included in the background slot. We also use the training camera poses of uORF to re-render our scenes in blender. Then uORF can obtain some foreground objects for some scenes, shown in Figure~\ref{fig:uorf}, which can roughly separate the blue chair from the whole scene while the brown desk still fails to be separated. But the segmented chair is unclear and blurry in the uORF results. For some other scenes, uORF still cannot properly segment the scene even after supplying the given camera poses. 

For 2D unsupervised model, we compare with slot attention~\cite{SlotAttention}. We adopt their model which is trained on CLEVR dataset~\cite{Johnson_2017_CVPR}, and then fine-tune the pretrained model on each of our scenes respectively. In most of the scenes, slot attention can segment some colors. However, without adequate 3D consideration or supervision, typically~\cite{SlotAttention} cannot cleanly separate the background, especially those scenes with textured background. Moreover, as our dataset is quite diverse, slot attention cannot achieve a very good reconstruction on par with the results produced by our model.


\subsubsection{3D Scene Editing}

\begin{figure*}[t]
\centering
    \includegraphics[width=0.7\linewidth]{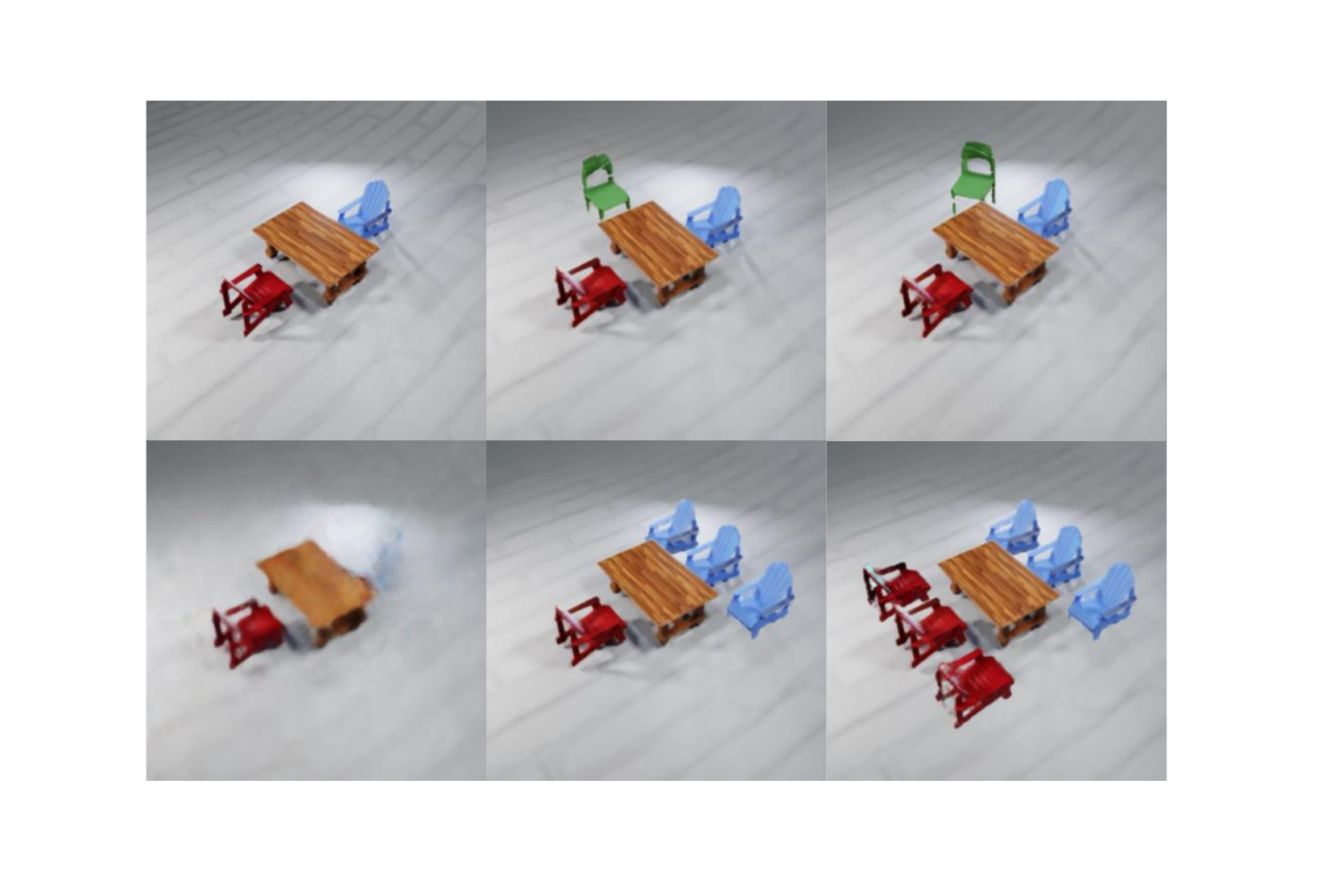}
    \vspace{-0.2in}
    \caption{\textbf{Results of object insertion, deletion and transformation.} The figure in the top left corner is the original scene image. The other two subfigures in the top row show insertions of a green chair. The bottom left corner shows deletion of the chairs. The remaining two show the results of adding more objects. }
    \label{fig:edit}
    \vspace{-0.2in}
\end{figure*}

After obtaining the masks and the individual object \nerfs, we can perform 3D scene editing. Generally, we can implement any rigid transformation with transformation matrices on the respective object \nerfs. Moreover, we can perform object insertion and removal by operating on the \nerfs separated by our method. While rendering the composite scene, by integrating the colors of all objects with the density contribution of each object, we can coherently handle occlusion after adding more objects to the scene. Figure~\ref{fig:edit} shows the results of multiple objects insertion. Given the \nerfs of the objects and a scene, we use the camera pose in the scene space to verify our editing results. Specifically, we use the RGB loss of the out-masked region of the images to get the NeRF of the scene after removing some objects. To translate an object in the scene, we translate the camera position for rendering. For example, suppose we want to move the object to $\mathbf p$. Let the pinhole camera be at $\bo$ and the object is  at the origin. Then we move the camera position $\bo' =\bo -\mathbf p$, where $\bo'$ is the new camera position. We sample points from $\bo'$ along the original directions to obtain the results after moving objects.

\subsubsection{Results on Real Scenes}

\begin{figure}[h]
    \includegraphics[width=1\linewidth]{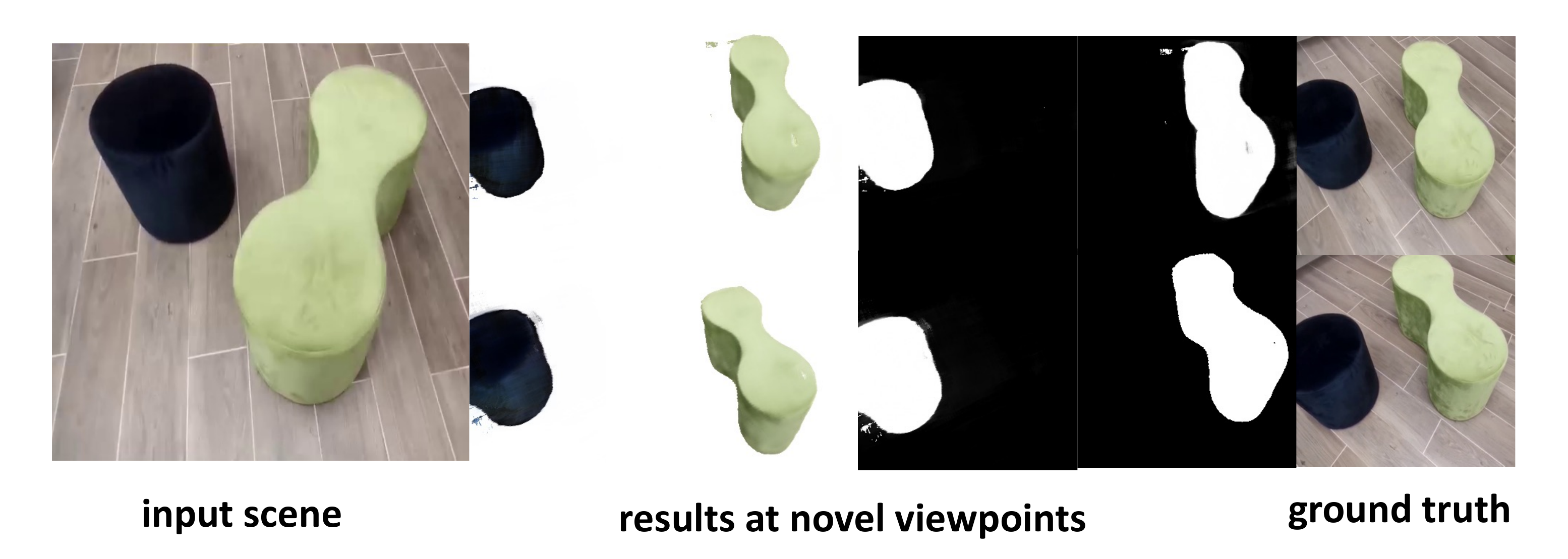}
    \caption{\textbf{ONeRF of real scene.} Here we show the result tested on a real scene. We visualize the result rendered at two novel viewpoints.}
    \label{fig:real_scene}
\end{figure}
Figure~\ref{fig:real_scene} shows the result tested on a real scene. Although the original images from real scene contain more noise due to complex real shading and non-homogeneous color distribution on a single object, our method is quite robust to such  noise, and achieves a reasonable separation in this real-world scenario.

\subsection{Quantitative Evaluation}

\begin{table}[t]
\footnotesize
\newcommand{\xpm}[1]{{\tiny$\pm#1$}}
\centering 
\begin{tabular}{lccc}
\toprule
  Method        & Mask IoU $\uparrow$ & Depth Error $\downarrow$ & PSNR $\uparrow$ \\ \midrule
  Ours          & $\mathbf{0.86}$ \xpm{0.0} & $\mathbf{0.002}$ \xpm{0.0} & $\mathbf{30.23}$ \xpm{0.0} \\ \midrule

  uORF~\cite{yu2021} & $\mathbf{0.50}$ \xpm{0.0} & - & $\mathbf{23.80}$ \xpm{0.0} \\ \midrule
  Slot attention~\cite{SlotAttention}    & $\mathbf{0.40}$ \xpm{0.0} & - & $\mathbf{18.99}$ \xpm{0.0} \\
\bottomrule
\end{tabular}
\caption{\textbf{Quantitative comparison.} We evaluate the segmented 3D objects on various metrics by comparing to the ground-truth renderings on from novel unseen viewpoints. For depth error, we normalize the depth between 0 and 1 and find the average mean square error. For uORF, it can only segment 2 objects out of all the 8 so we take the average on these two objects. The depth is used to estimate the 3D position, which is not applicable for~\cite{yu2021} and~\cite{SlotAttention}.}
\label{tab:compare}
\vspace{-0.5em}
\end{table}

\cref{tab:compare} shows the quantitative comparison between our method and the two baseline methods, where we can see that in our moderately complex and diverse scenes scenario, our method produces the most stable performance.

\subsection{Ablation Studies}

\paragraph{Effect of mask refinement.} 

\begin{figure}[t]
    \includegraphics[width=1\linewidth]{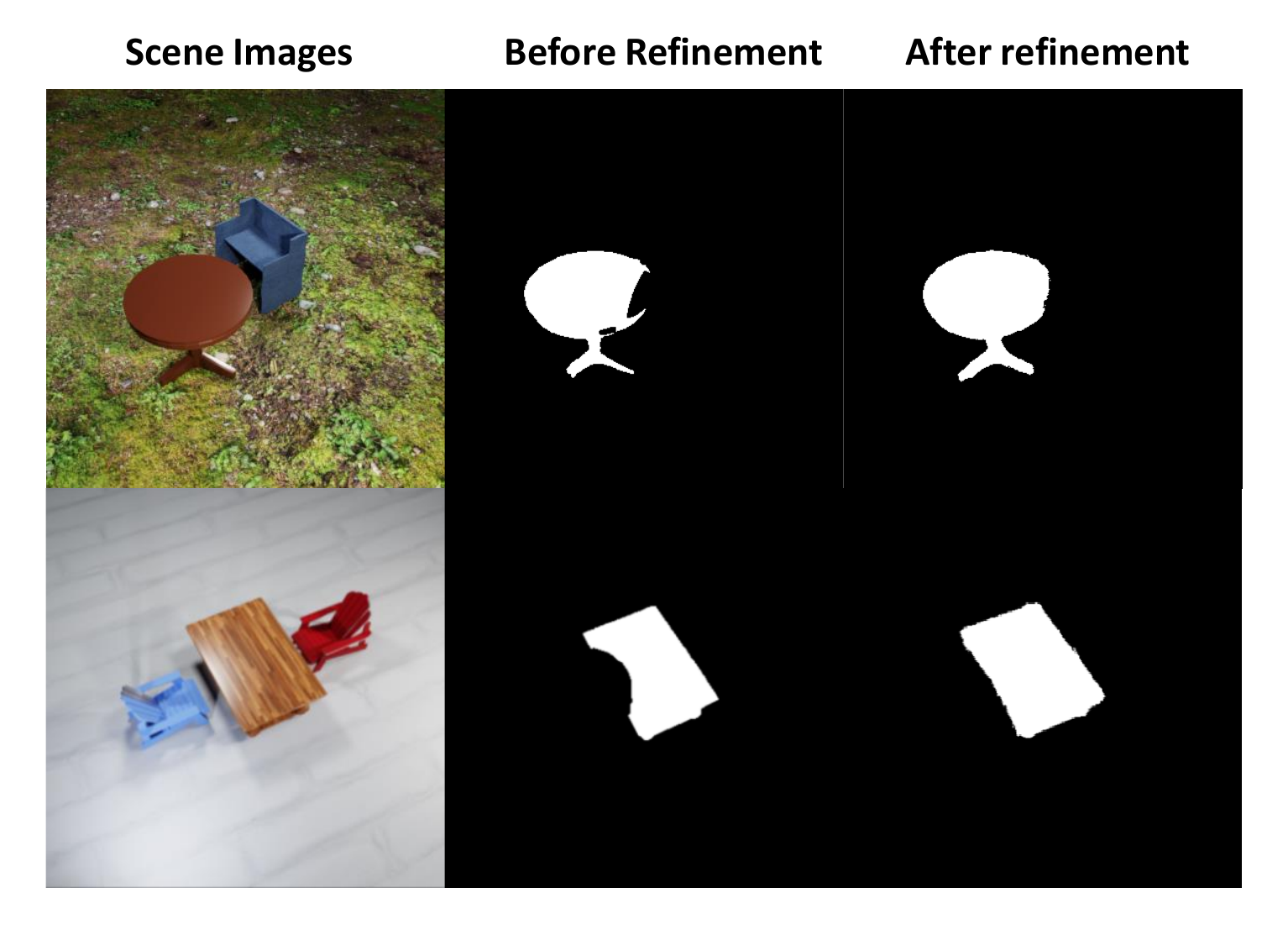}
    \caption{\textbf{Masks Comparison.} Here are the masks of the two tables before and after refinement.  }
    \label{fig:mask}
\end{figure}

We compare the masks before and after our refinement using intersection over union. The errors and figures for comparison are shown in Table ~\ref{tab:mask} and Figure~\ref{fig:mask}. The refinement step can complete both tables. 

\begin{table}[t]
\footnotesize
\newcommand{\xpm}[1]{{\tiny$\pm#1$}}
\centering 
\begin{tabular}{lccc}

\toprule
  Method    & before refinement     & before refinement   & \\ \midrule
  Mask IoU $\uparrow$  & $\mathbf{0.86}$ & $\mathbf{0.76}$  \\ 

\bottomrule



\end{tabular}
\caption{\textbf{Masks comparison.} We evaluate the IoU errors of all objects over the 8 scenes before and after the refinement step.}
\label{tab:mask}
\vspace{-0.5em}
\end{table}


\paragraph{Effect of object segmentation network.}

\begin{figure*}[h]
\centering
    \includegraphics[width=1\linewidth]{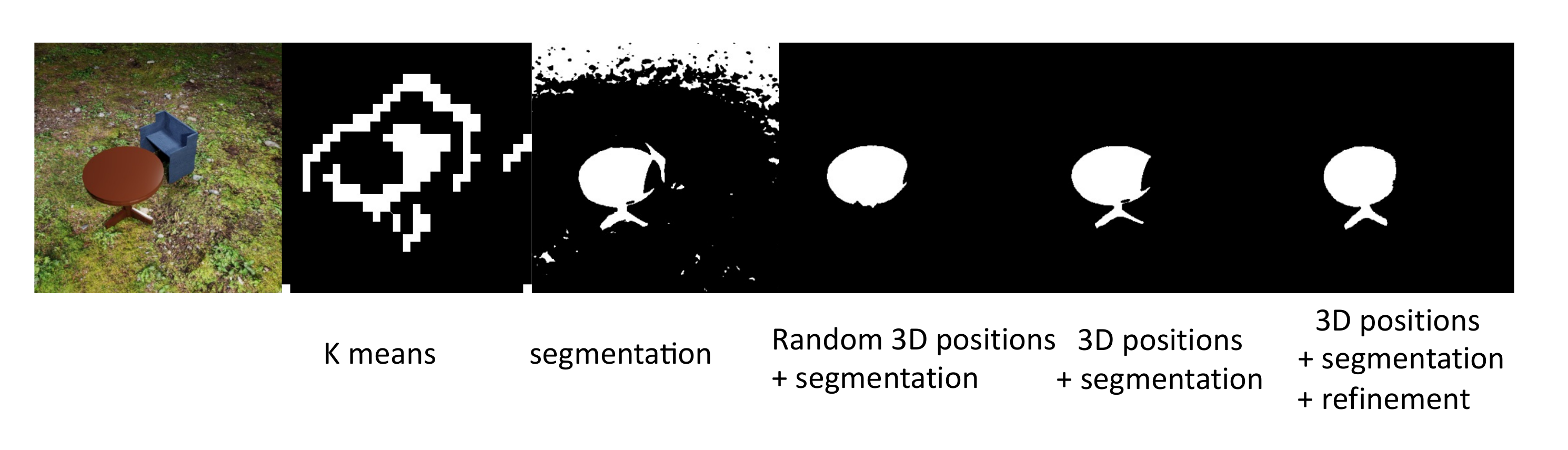}
    \vspace{-0.2in}
    \caption{\textbf{Ablation study.} Here we visualize the results of the ablation study. From left to right: the ground truth image at novel view, result without segmentation model, result without foreground-background separator, result without correct 3D information, result without refinement and result produced by our full model.}
    \label{fig:ablation}
\end{figure*}

As shown in the second column of Figure~\ref{fig:ablation}, just simple $K$-means clustering cannot replace our more advanced segmentation model, since the former cannot precisely distinguish coherent object  boundary and is very vulnerable to noise.

\paragraph{Effect of coarse clustering.}

We use the VGG feature to distinguish the background and foreground objects, and 3D positions to select foreground objects. This method can roughly choose foreground object automatically while removing a large portion of background regions in the images, which helps the subsequent object segmentation. 

During our testing, we find that with the same initial $K$-Means clustering parameters, adding 3D positions can cluster 
the foreground objects to the specific class where we use 3D information. Additionally, we remove the coarse clustering step, i.e., directly apply the unsupervised segmentation method~\cite{9151332} to the same input images in our testing. Then we replace the predicted 3D positions with random noise in the first step to check whether the positions effects. Both results are shown in Figure~\ref{fig:ablation}.

\section{Conclusions}
This paper presents the first significant step toward segmenting multiple objects \nerfs from multiple views of the underlying scene. Compared with single-image methods, with the 3D from \nerfs, O\nerfs can alternately optimize 2D masks and 3D geometry so that the former can be used to delineate cluttered scenes into coherent object \nerfs, which can then be used to enable a range of 3D scene editing tasks, allowing users to readily insert 3D  objects, delete 3D objects, and modify object geometry in addition to novel view synthesis. In the future, we will investigate \nerf completion when the pertinent occluded parts of the object were not photographed by any of the  input views. This  is achieved by detecting the underlying object symmetry, which would not have been possible without segmenting the scene into individual object \nerfs.

{\small
\bibliographystyle{ieee_fullname}
\bibliography{ref}
}

\end{document}


\title{\LaTeX\ Author Guidelines for \confName~Proceedings}

\author{First Author\\
Institution1\\
Institution1 address\\
{\tt\small firstauthor@i1.org}
\and
Second Author\\
Institution2\\
First line of institution2 address\\
{\tt\small secondauthor@i2.org}
}
\maketitle

{\small
\bibliographystyle{ieee_fullname}
\bibliography{ref}
}